%% file: paper3.tex
\documentclass[11pt]{article}
\usepackage[final]{acl/acl}

\usepackage{times}
\usepackage{latexsym}
\usepackage[T1]{fontenc}
\usepackage{microtype}
\usepackage{inconsolata}
\usepackage{booktabs}
\usepackage{amsmath,amssymb}
\usepackage{graphicx}
\usepackage{xurl}
\usepackage{placeins}
\usepackage{float}
\usepackage{algpseudocode}
\usepackage{parskip}

\newfloat{algorithm}{t}{loa}
\floatname{algorithm}{Algorithm}

\newcommand{\code}[1]{\texttt{#1}}

\title{Learning When to Commit from Partial Speech for End-to-End Simultaneous Speech Translation}
\author{Hieu Hoang \\
  Microsoft\\
  \And
  Amittai Axelrod}

\begin{document}
\maketitle

\input{0_abstract}
\input{1_intro}
\input{2_related_work}
\input{3_experimental_setup}
\input{4_results}
\input{5_conclusion}

\input{6_limitations}

\FloatBarrier

\bibliography{references}

\appendix
\input{7_appendix}

\end{document}

%% file: 0_abstract.tex
\begin{abstract}
Simultaneous speech translation must emit useful target text before the source is complete while preserving every committed token. We adapt a full-utterance speech language model using prefix supervision derived from its own complete- and partial-waveform translations, requiring neither transcripts nor human translations. We compare single-turn forced-prefix and multi-turn append-only decoding, use a confidence threshold to control the inference-time quality--latency trade-off, and vary the density of training prefixes with a separate synthesis margin. On FLEURS and CoVoST\,2 in three language directions, prefix training improves quality--latency frontiers over the unadapted model, and confidence provides the broadest consistently competitive operating range. Multi-turn decoding is generally stronger at low latency; under multi-turn training, commit-calibration error falls by 63--68\% overall and 68--80\% at early prefixes, whereas single-turn training provides only modest overall calibration gains and no early-prefix improvement. A small synthesis margin sometimes extends the frontier to lower latency, particularly on shorter utterances, while a larger margin degrades translation quality and calibration. Prefix adaptation therefore improves simultaneous speech translation, especially under multi-turn append-only decoding, while synthesis density introduces a non-monotonic quality--latency trade-off.
\end{abstract}

%% file: 1_intro.tex
\section{Introduction}

Simultaneous speech translation begins producing a target-language rendering while the speaker is still talking. The system must therefore act on an evolving acoustic prefix, decide whether the available evidence licenses new target content, and preserve everything already shown or spoken. Waiting for the complete utterance maximizes context but incurs sentence-level latency; translating every incoming block minimizes delay but risks omissions, reordering errors, and unstable output.

These decisions are especially consequential for direct speech language models. Unlike a text translation component behind streaming ASR, an end-to-end model receives no explicit word boundary or finalized transcript. Acoustic evidence arrives in time, source words may straddle blocks, and both recognition and translation uncertainty are represented inside one model. Yet modern speech language models are generally optimized for complete inputs and ordinary assistant responses, not for repeatedly extending an irrevocable translation from partial audio.

Prior work addressed the analogous text problem by making a sentence-trained language model prefix-aware \citep{hoang-axelrod-post:2026:wmt}. Its central construction uses the model's own full-input translation as an anchor, derives targets from translations of source prefixes, and replays previously committed target text as a forced prefix. A confidence threshold then lets the model vary its lag dynamically rather than following the same read--write schedule for every sentence. Applying this principle to direct speech is not a direct modality substitution: acoustic inputs form much longer encoded sequences, arrive at model-dictated block boundaries, lack explicit lexical boundaries, and couple recognition and translation uncertainty. Efficient streaming must therefore preserve prior audio computation while representing each commit decision in a causal history of audio blocks, waits, and target emissions.

Our starting point during training is a self-teacher speech language model. We ask the model to translate complete utterances and fixed-duration prefixes of the same waveform. Agreement with the full-audio translation defines how much target text can be supervised for each prefix audio. We further introduce a synthesis margin that allows the full-audio translation to rescue plausible target tokens during prefix construction, explicitly controlling the density of intermediate supervision. This training-time margin is distinct from the inference-time confidence threshold that controls when the trained model emits. The same English audio supports all target languages, and no human translation labels are required to construct the chains.

Speech input also exposes a representation choice that is less consequential in the text-prefix setting. Audio inputs are much longer than text inputs, making reuse of prior computation especially important. In the \emph{single-turn forced-prefix decoding} protocol, all observed audio precedes an open assistant turn containing the committed target as a forced prefix. Subsequent audio blocks are appended to the prompt. Therefore, the KV cache is retained only through the audio boundary; the full committed target must be replayed on a temporary extension before each new delta is generated, after which those target states are discarded. For long sequences, this replay can become computationally expensive, motivating the multi-turn append-only approach.

In the \emph{multi-turn append-only decoding} protocol, each new audio block and target delta is appended to one causal sequence, so all existing key--value states can be reused. We compare how these representations affect the quality--latency trade-off under the same training and evaluation data.

At inference, a confidence threshold compares continuation and stopping evidence using only the observed audio and committed target. Sweeping this threshold gives us a quality--latency control, which we compare with fixed wait-$k$ and target-suffix deletion. 

This work makes four contributions. First, it constructs target-prefix supervision directly from partial waveforms, without transcripts or human translations. Second, it introduces a synthesis margin that controls how readily the full-audio self-translation can extend intermediate targets, providing a controlled intervention on supervision density. Third, it formulates and compares single-turn and multi-turn prompting protocols for streaming audio. Fourth, it shows that prefix adaptation improves quality--latency frontiers under both protocols and is associated with substantially better commit calibration under multi-turn decoding.

%% file: 2_related_work.tex
\section{Related Work}

\paragraph{Simultaneous translation policies.} Simultaneous translation is commonly formulated as a sequence of read and write decisions. 

Learned policies use reinforcement learning or supervision from oracle actions \citep{gu-etal-2017-learning,zheng-etal-2020-simultaneous}, while wait-$k$ provides a simple fixed schedule by reading an initial source prefix and then alternating reads and writes \citep{ma-etal-2019-stacl}. Fixed policies expose a clear latency knob but cannot adapt their delay to the uncertainty of an individual sentence. Our confidence mechanism instead lets the model decide how much target text to append after the arrival of each audio block, with a threshold controlling the required evidence for commitment.

\paragraph{Streaming speech translation.} Speech translation systems may cascade streaming ASR into text MT or translate speech directly \citep{bentivogli-etal-2021-cascade}. Cascades benefit from mature components but expose the translation model to changing recognition hypotheses. Direct systems avoid this interface, although they must learn acoustic recognition, translation, and scheduling jointly. Streaming transducers explicitly alternate source consumption and target generation \citep{cho-esipova-2016-can,huang-etal-2020-simultaneous}; retranslation systems repeatedly process the available prefix with a sentence model \citep{niehues-etal-2016-dynamic}. We retain the flexibility of prefix retranslation while constraining the target to be append-only, and we apply it directly to waveform prefixes rather than reference or ASR transcripts.

\paragraph{Output stability and prefix supervision.} Retranslation can revise earlier words, producing flicker. Prior work reduces revision through target-suffix deletion \citep{arivazhagan-etal-2019-retranslation}, forced translation \citep{polak-etal-2022-cuni}, or agreement between consecutive hypotheses \citep{liu-etal-2020-low}. Stable-prefix ideas have therefore primarily served as inference-time commit rules in retranslation systems \citep{niehues-etal-2016-dynamic,niehues-etal-2018-low}. Self-training on partial inputs can also make a model more robust to incomplete context \citep{sen-etal-2023-self}. We use agreement with the model's unconstrained full-audio translation to create training targets and preserve the committed target at inference time, making our approach output stable by construction.

\paragraph{Speech language models.} Recent work increasingly connects speech encoders to general-purpose language models and studies their use in end-to-end speech applications \citep{yang-etal-2025-large-language}. Multi-turn speech instruction data \citep{ouyang-etal-2025-infinisst,wang-etal-2025-conversational}, LLM-oriented simultaneous-translation frameworks \citep{agostinelli-etal-2024-simul}, and interleaved source--target training under latency constraints \citep{fu-etal-2025-llms} all move toward models that reason and generate across modalities. Our focus is narrower: adapting one offline speech language model to produce an irrevocable text translation from growing audio, while holding the full-audio self-translation fixed as the supervision anchor.

\paragraph{Confidence and calibration.} Dynamic commitment requires a score whose meaning remains useful across source positions. Neural sequence models can be miscalibrated, including in their end-of-sequence probabilities, and calibration often degrades under distribution shift \citep{kumar-sarawagi-2019-calibration,ovadia-etal-2019-trust}. 

Here, applying a model tuned for complete utterances to partial speech introduces such a train--inference mismatch, motivating direct evaluation of commit calibration. Given the next-token distribution for streaming context $s$, let
\begin{equation}\label{eq:commit-probabilities}
\begin{aligned}
p_{\mathrm{cont}}&=\max_{v\notin\mathcal{E}}p(v\mid s), &
p_{\mathrm{EOS}}&=\max_{e\in\mathcal{E}}p(e\mid s),
\end{aligned}
\end{equation}
where $\mathcal{E}$ is the set of end-of-sequence tokens. At each non-final boundary, the deployment rule is
\begin{equation}\label{eq:commit-deployment-rule}
\begin{aligned}
\rho &= p_{\mathrm{cont}}-p_{\mathrm{EOS}},\\
a(\rho;\tau) &=
\begin{cases}
\textsc{commit}, & \rho\geq\tau,\\
\textsc{stop}, & \rho<\tau.
\end{cases}
\end{aligned}
\end{equation}


%% file: 3_experimental_setup.tex
\section{Experimental Setup}
\label{sec:experimental-setup}

\subsection{Task and Languages}

We study direct English-speech-to-text translation in three directions: EN$\rightarrow$DE, EN$\rightarrow$JA, and EN$\rightarrow$ZH. 
At streaming step $i$, the system observes waveform prefix $a_{\leq t_i}$ and an irrevocable target translation prefix $c_{i-1}$. The system then either waits or appends new output $\Delta_i$ to the target translation. 
The visible translation is $c_i=c_{i-1}\circ\Delta_i$. Our LLM's native audio implementation dictates a fixed $2$\,s block cadence, so source time advances in $2$\,s waveform blocks, with a final shorter block when necessary. We call the model-visible context at boundary $i$ the \emph{commit-decision context}: it comprises the observed audio, previously committed target, and protocol-specific causal history. Single-turn decoding reconstructs this context from cumulative audio and a forced target prefix, whereas multi-turn decoding retains the interleaving sequence of audio blocks and target deltas.

\subsection{Translation Prefix Synthesis}
\label{sec:speech-prefixes}

Let $g$ denote the offline speech language model, let $f=g(a)$ be its translation of the complete utterance $a$, and let $h_i=g(a_{\leq t_i})$ be its unconstrained translation of audio prefix $i$. If $\ell_i=|\operatorname{lcp}(h_i,f)|$, the supervised commitment at boundary $i$ is
\begin{equation}
  c_i=f_{1:m_i},\qquad m_i=\max_{j\leq i}\ell_j.
\end{equation}
We measure target agreement in words for German and characters for Japanese and Chinese. The running maximum makes the chain monotone even when a later prefix decode is less stable than an earlier one. The complete-audio row always supplies $f$. Algorithm~\ref{alg:stable-prefix-synthesis} in the Appendix gives the complete synthesis procedure.

\subsubsection{Synthesis Margin}

Merging the idea of translation prefix synthesis and confidence-based token selection of Equations~\ref{eq:commit-probabilities} and~\ref{eq:commit-deployment-rule}, we apply a confidence-based margin to token selection during target-prefix creation. Let $r$ be the next token in the self-teacher's full-audio translation and $x$ the greedy next token given the current partial-audio prefix. The next synthesis action is
\begin{equation}
\tilde{x}=\begin{cases}
r, & x=r \text{ or } p(r)+\delta>p(x),\\
\textsc{stop}, & \text{otherwise}.
\end{cases}
\end{equation}
We call $\delta$ the \emph{synthesis margin}. The setting $\delta=0$ recovers the original prefix algorithm, while increasing $\delta$ permits the full-audio reference to rescue a less probable token. This can lengthen prefix supervision at the risk of lower quality translation.

\subsection{Decoding Protocols}

\paragraph{Single-turn forced-prefix decoding} renders each commit-decision context as a single multimodal user prompt containing a translation instruction and the observed waveform prefix. The committed target is appended as an open assistant prefix, and the model is trained to generate only the new continuation (and the closing token). The user prompt indicates the final boundary with an end-of-audio marker. This is analogous to the successful prefix model for text-based simultaneous translation in \citep{hoang-axelrod-post:2026:wmt}.

\paragraph{Multi-turn append-only decoding} converts cumulative commitments $c_1,\ldots,c_K$ into append-only deltas and forms one causal record that alternates audio blocks and assistant deltas. Cross-entropy is applied to each target delta (and its closing token). The final sequence includes an end-of-audio marker before its last target delta.

Both protocols are flicker-free by construction and permit the prompt and encoded audio prefix to remain cached. Multi-turn append-only decoding further reduces redundant computation by reusing previous assistant deltas. Appendix~\ref{sec:multi-turn-template} gives the complete prompt templates for both protocols.

\subsection{Model and Training}

The base model and self-teacher are both Qwen2.5-Omni-7B \citep{xu-etal-2025-qwen25-omni}. LoRA adapters \citep{hu-etal-2022-lora} are applied to the text decoder with rank 32, scale 64, and dropout $0.05$. Training begins from the same base model. Optimization uses AdamW \citep{kingma-ba-2015-adam}, a constant learning rate of $5\times10^{-6}$, and an effective batch size of 64. We retain checkpoints every 500 updates through a maximum of 6,000 updates. Other details of the training recipe can be found in Appendix~\ref{sec:reproducibility}.

\paragraph{Single-turn forced-prefix} is trained on two equally sampled tasks, both with an empty committed target: a partial-audio example targeting prefix translation at one retained $2$\,s boundary, and a full-audio example targeting complete translations. Loss is applied only to the target text and closing token.

\paragraph{Multi-turn append-only} Each full utterance forms one causal record containing all $2$\,s audio blocks and target deltas in order. An assistant turn with no target text supervises a wait. Loss covers every assistant delta and closing token in one forward pass.

\subsection{Data}

We train on LibriSpeech-100 and LibriSpeech-360 \citep{panayotov-etal-2015-librispeech} and Common Voice \citep{ardila-etal-2020-common}, validate on a mixed set of utterances held out from the training data, and test on FLEURS \citep{conneau-etal-2022-fleurs} and CoVoST~2 \citep{wang-etal-2020-covost2}. Appendix Table~\ref{tab:data-splits} summarizes the data. We reserve every 50th training utterance, using the first 400 reserved utterances for mixed validation and the remainder as a holdout. Utterance selection and $2$\,s prefix boundaries are shared across decoding protocols. Common Voice training candidates are decontaminated against CoVoST~2 by identifier, audio basename, normalized transcript, and audio hash.

An analysis of the full-audio found that EN$\rightarrow$ZH translations often contained conversational boilerplate, particularly in the Common Voice subset. For example, a translation of ``a lot of money,'' might appear as ``a lot of money \emph{; if there is any more audio content, you can continue telling me.}''. 

We conservatively remove trailing clauses that invite the user to provide more audio or request further translation help, then truncate every prefix translation accordingly while preserving monotonicity. This changes 92,039 of 139,684 chains and quarantines eight empty results. Appendix~\ref{sec:zh-anchor-cleaning} specifies the rule and reports the cleanup audit.

Table~\ref{tab:synthesis-prefix-length} confirms that the mean target-prefix length increases monotonically with $\delta$ for every language and source corpus when synthesis margins are applied.
\begin{table}[t]
\centering
\scriptsize
\caption{Effect of the synthesis margin on the training data, measured by the mean cumulative target-prefix length at each non-final $2$\,s audio block.}
\label{tab:synthesis-prefix-length}
\setlength{\tabcolsep}{4pt}
\begin{tabular}{@{}llrrr@{}}
\toprule
Language & Training dataset & $\delta=0$ & $\delta=0.2$ & $\delta=0.5$ \\
\midrule
EN$\rightarrow$DE & LibriSpeech-100 & 10.2 & 15.9 & 18.9 \\
 & LibriSpeech-360 & 10.2 & 16.0 & 18.9 \\
 & Common Voice & 4.2 & 5.5 & 6.5 \\
\addlinespace[2pt]
EN$\rightarrow$JA & LibriSpeech-100 & 16.7 & 35.4 & 46.4 \\
 & LibriSpeech-360 & 16.6 & 35.3 & 46.1 \\
 & Common Voice & 9.5 & 13.9 & 20.5 \\
\addlinespace[2pt]
EN$\rightarrow$ZH & LibriSpeech-100 & 14.1 & 25.5 & 35.5 \\
 & LibriSpeech-360 & 14.1 & 25.3 & 35.3 \\
 & Common Voice & 7.4 & 10.2 & 13.7 \\
\bottomrule
\end{tabular}
\end{table}

\subsection{Validation and Checkpoint Selection}

\paragraph{Single-turn forced-prefix decoding.} Every checkpoint is first scored on at most 400 utterances from the deterministic one-in-50 LibriSpeech-100 chain holdout. Direct validation computes COMET for stable partial-audio targets and complete-audio translations, and ranks checkpoints by their equal-weight mean. The top two candidates undergo autoregressive FLEURS-development rollouts over $\tau\in\{-0.8,-0.6,-0.3,0.0,0.3,0.6,0.8,0.9,1.0\}$. A full-audio quality gate requires the candidate's $\tau{=}1.0$ COMET to remain within $0.5$ of the base self-teacher. Among eligible candidates, mean attainable COMET over a shared latency-budget grid determines the selection; the direct score breaks ties.

\paragraph{Multi-turn append-only decoding.} Controlled validation on the same deterministic holdout evaluates productive partial contexts and final contexts with gold causal history. The productive-partial COMET champion and WAIT-balanced-accuracy champion form the shortlist. These candidates then undergo append-only autoregressive development rollouts over the same nine confidence thresholds, and mean attainable COMET over their shared latency support determines the selected checkpoint.

\subsection{Inference Mechanisms and Metrics}

All inference mechanisms receive the same sequence of audio blocks and preserve the previously committed target. The confidence mechanism compares the highest non-EOS continuation probability with the EOS probability and emits only when their margin exceeds threshold $\tau$; sweeping $\tau$ traces a model's quality--latency frontier. We compare it with wait-$k$ and target-suffix deletion \citep{ma-etal-2019-stacl,arivazhagan-etal-2019-retranslation}. Wait-$k$ suppresses writing for a fixed number of initial $2$\,s blocks, while suffix deletion withholds a fixed number of trailing target units from each non-final delta.

Translation quality is measured with COMET-22-DA \citep{rei-etal-2022-comet}. Latency is measured with Average Lagging in seconds, using audio blocks rather than source-token position, as defined in Appendix~\ref{sec:reproducibility}, Equation~\ref{eq:average-lagging}. 

We evaluate the selected single-turn forced-prefix decoding and multi-turn append-only decoding models on FLEURS primarily, and on CoVoST~2 at the end. We report each complete frontier rather than a single operating point, and include the complete-audio translation as the offline endpoint.

Commit calibration is evaluated on identical oracle commit-decision contexts for every model using expected calibration error with utterance-cluster bootstrap intervals. The oracle is derived from the base model's running-max prefix chain so that models are compared on the same continue-versus-stop decisions.

%% file: 4_results.tex
\section{Results}
\label{sec:results}

\subsection{Single-turn Forced-prefix Decoding}

\begin{figure*}[t]
\centering
\textbf{FLEURS}\\[1pt]
\includegraphics[width=0.72\linewidth]{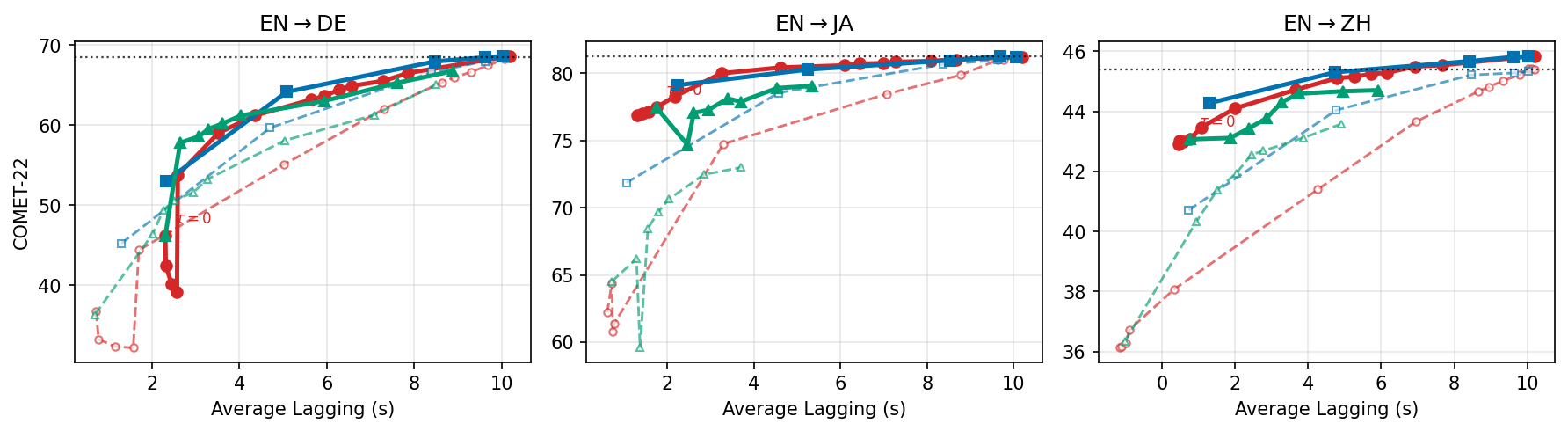}\\
\vspace{2pt}
\includegraphics[width=0.70\textwidth]{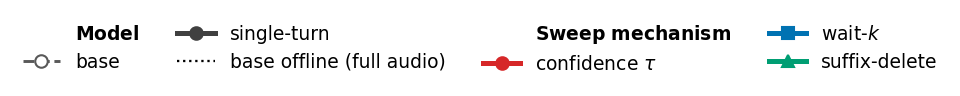}
\caption{Quality--latency frontiers for the base and canonical \code{single-turn} models under confidence, wait-$k$, and suffix-deletion controls.}
\label{fig:paper3-speech-staging-trained-controls}
\end{figure*}

Figure~\ref{fig:paper3-speech-staging-trained-controls} compares the three inference mechanisms on the base and \code{single-turn} models, and Table~\ref{tab:paper3-speech-staging-auc} summarizes every frontier with attainable-envelope AUC, defined in Appendix~\ref{sec:reproducibility}, Equation~\ref{eq:attainable-envelope-auc}.

The trained model consistently outperforms the base model across all language directions. Confidence margin gives the highest or tied trained-model AUC in all three language directions. It can occasionally be matched by the other mechanisms, but its range exceeds all others. 

\begin{table}[!t]
\centering
\scriptsize
\caption{AUC for the base and single-turn models in Figure~\ref{fig:paper3-speech-staging-trained-controls} over each panel's shared latency interval.}
\label{tab:paper3-speech-staging-auc}
\setlength{\tabcolsep}{5pt}
\begin{tabular}{@{}llrrr@{}}
\toprule
Direction & Model & wait-$k$ & suf.\,del. & confidence \\
\midrule
\multicolumn{5}{@{}l}{\textbf{FLEURS}} \\
EN$\rightarrow$DE & base & 54.0 & 54.8 & 52.2 \\
 & trained & 56.0 & \textbf{57.0} & 57.0 \\
EN$\rightarrow$JA & base & 75.6 & 71.6 & 73.3 \\
 & trained & 76.6 & 76.3 & \textbf{78.4} \\
EN$\rightarrow$ZH & base & 41.7 & 41.7 & 40.5 \\
 & trained & 43.1 & 42.9 & \textbf{43.5} \\
\bottomrule
\end{tabular}
\end{table}

\subsection{Multi-turn Append-Only Decoding}

\begin{figure*}[t]
\centering
\textbf{FLEURS}\\[1pt]
\includegraphics[width=0.72\linewidth]{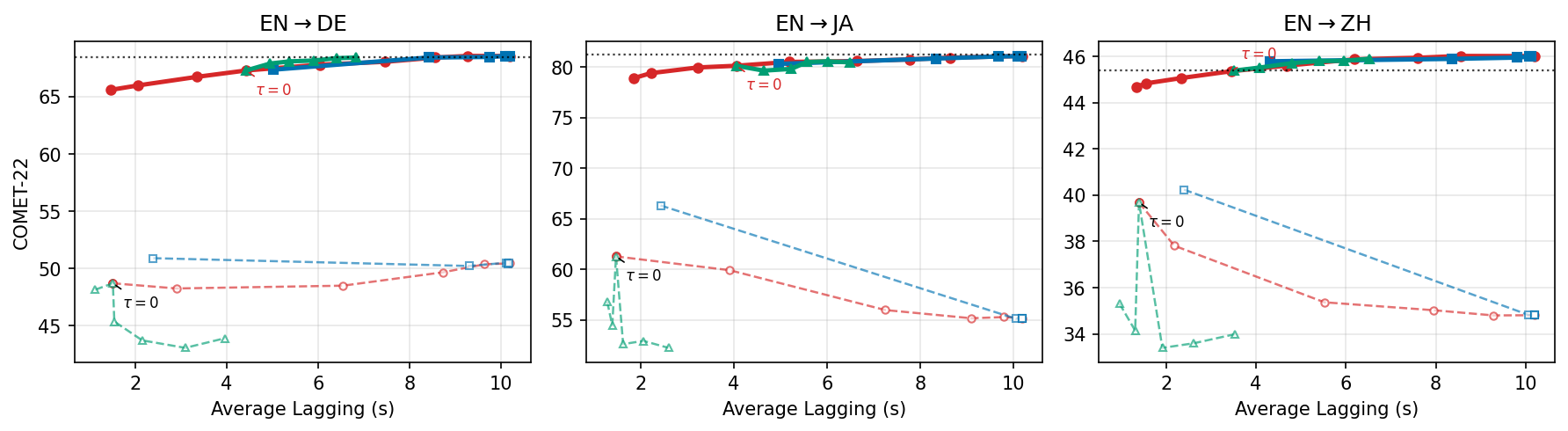}\\
\vspace{2pt}
\includegraphics[width=0.70\textwidth]{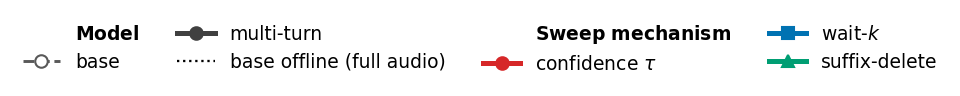}
\caption{Quality--latency frontiers for the base and canonical \code{multi-turn} models.}
\label{fig:paper3-speech-frontier-il-trained-controls}
\end{figure*}

Figure~\ref{fig:paper3-speech-frontier-il-trained-controls} and Table~\ref{tab:paper3-speech-frontier-il-auc} compare the three inference mechanisms on the base and \code{multi-turn} models. 

The base model degenerates under multi-turn incremental decoding, producing lower-quality translations whose quality can decrease even as more source context becomes available. This pattern is consistent with a train--inference mismatch: the base model was not trained to extend translations across interleaved audio and target turns. The trained model is better by far and, with the confidence mechanism, has the best latency range.

\begin{table}[H]
\centering
\scriptsize
\caption{AUC for the base and multi-turn models in Figure~\ref{fig:paper3-speech-frontier-il-trained-controls} over each panel's shared latency interval.}
\label{tab:paper3-speech-frontier-il-auc}
\setlength{\tabcolsep}{5pt}
\begin{tabular}{@{}llrrr@{}}
\toprule
Direction & Model & wait-$k$ & suf.\,del. & confidence \\
\midrule
\multicolumn{5}{@{}l}{\textbf{FLEURS}} \\
EN$\rightarrow$DE & base & 49.7 & 48.7 & 48.6 \\
 & trained & 57.2 & 59.0 & \textbf{66.4} \\
EN$\rightarrow$JA & base & 64.5 & 61.1 & 61.0 \\
 & trained & 68.8 & 71.5 & \textbf{78.4} \\
EN$\rightarrow$ZH & base & 39.2 & 39.5 & 39.4 \\
 & trained & 41.3 & 42.3 & \textbf{45.0} \\
\bottomrule
\end{tabular}
\end{table}

\subsection{Synthesis Margin}

\begin{figure*}[t]
\centering
\textbf{FLEURS}\\[1pt]
\includegraphics[width=0.72\linewidth]{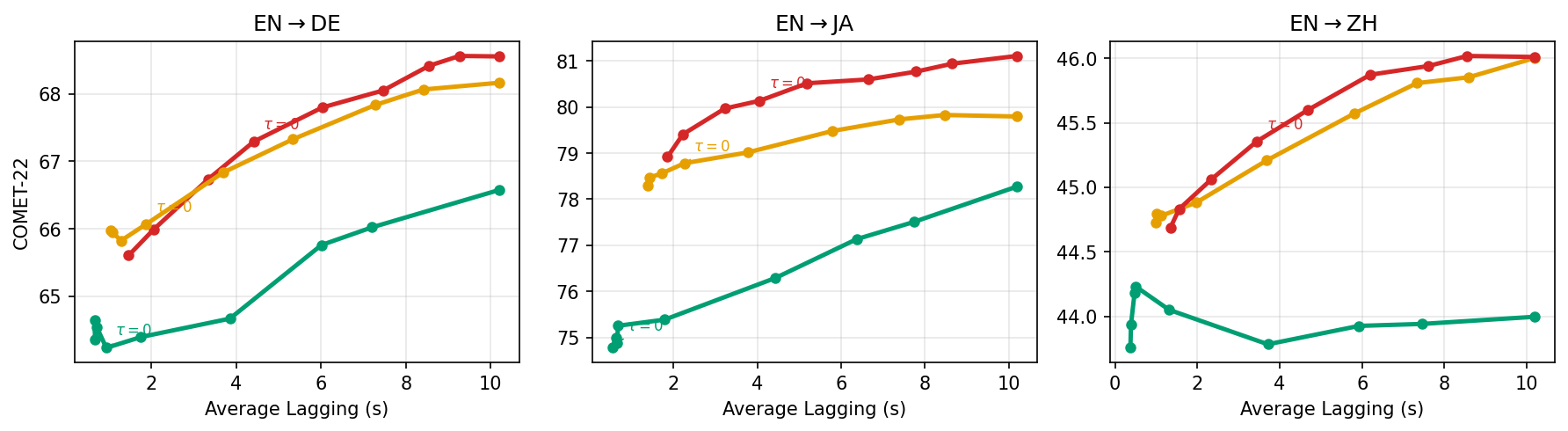}\\
\vspace{3pt}
\textbf{CoVoST\,2}\\[1pt]
\includegraphics[width=0.72\linewidth]{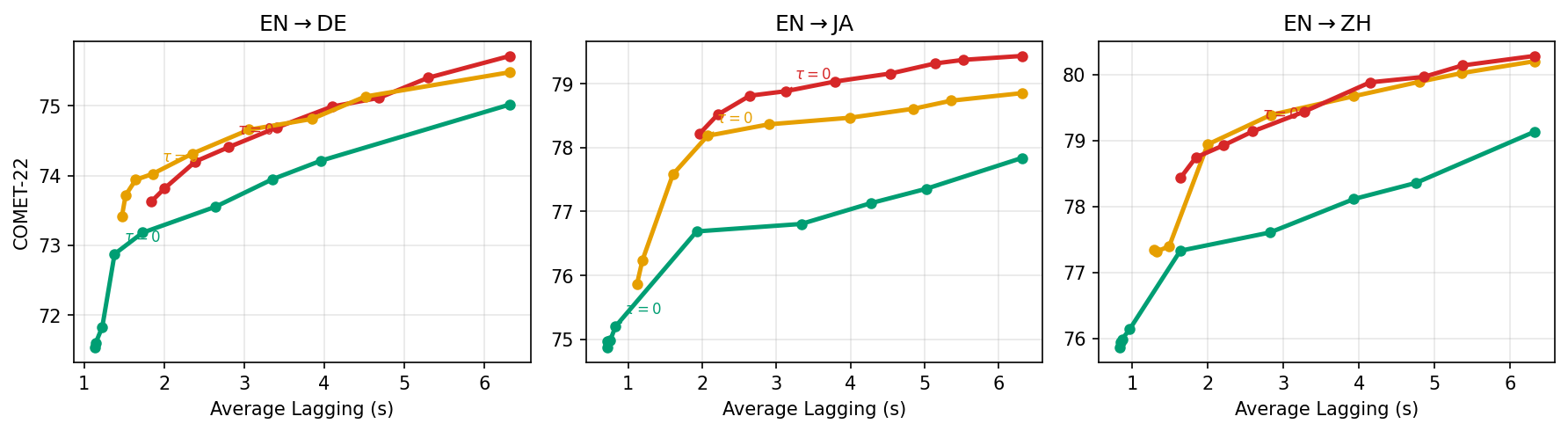}\\
\vspace{2pt}
\includegraphics[width=0.70\textwidth]{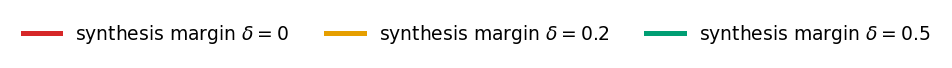}
\caption{Confidence-controlled multi-turn frontiers for models trained with synthesis margins.}
\label{fig:paper3-speech-synthesis-margin}
\end{figure*}

Figure~\ref{fig:paper3-speech-synthesis-margin} compares multi-turn models trained with synthesis margins. $\delta=0.5$ is uncompetitive and even degenerate for EN$\rightarrow$ZH when tested on FLEURS. $\delta=0.2$ matches or underperforms the original prefix model over most of the latency range. On the shorter CoVoST~2 utterances, however, its denser intermediate supervision extends the frontier to lower latency than the canonical multi-turn model can reach.

These results expose a trade-off between low latency and reliable translation. Training on corpora synthesized with a small margin of $\delta=0.2$, which commits target-prefix tokens more aggressively, can improve low-latency quality, but the gains vary by language pair and test set. At $\delta=0.5$, training produces a heavier tail of severe errors: some outputs drift into unsupported content or repetition, reducing translation fidelity.

\subsection{Single-turn vs. Multi-turn}

\begin{figure*}[t]
\centering
\textbf{FLEURS}\\[1pt]
\includegraphics[width=0.72\linewidth]{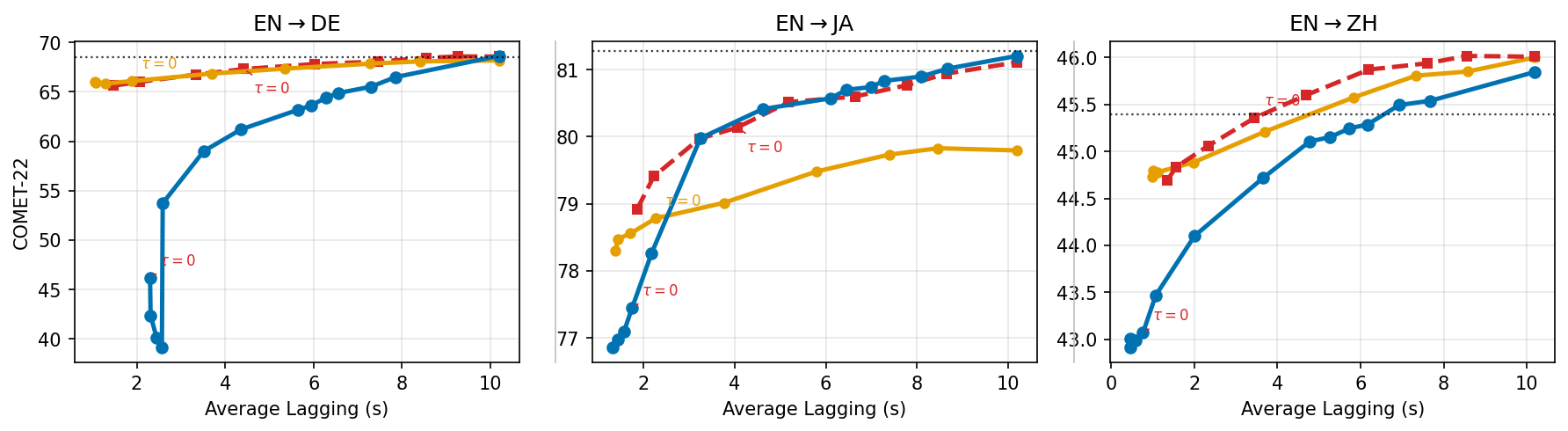}\\
\vspace{3pt}
\textbf{CoVoST\,2}\\[1pt]
\includegraphics[width=0.72\linewidth]{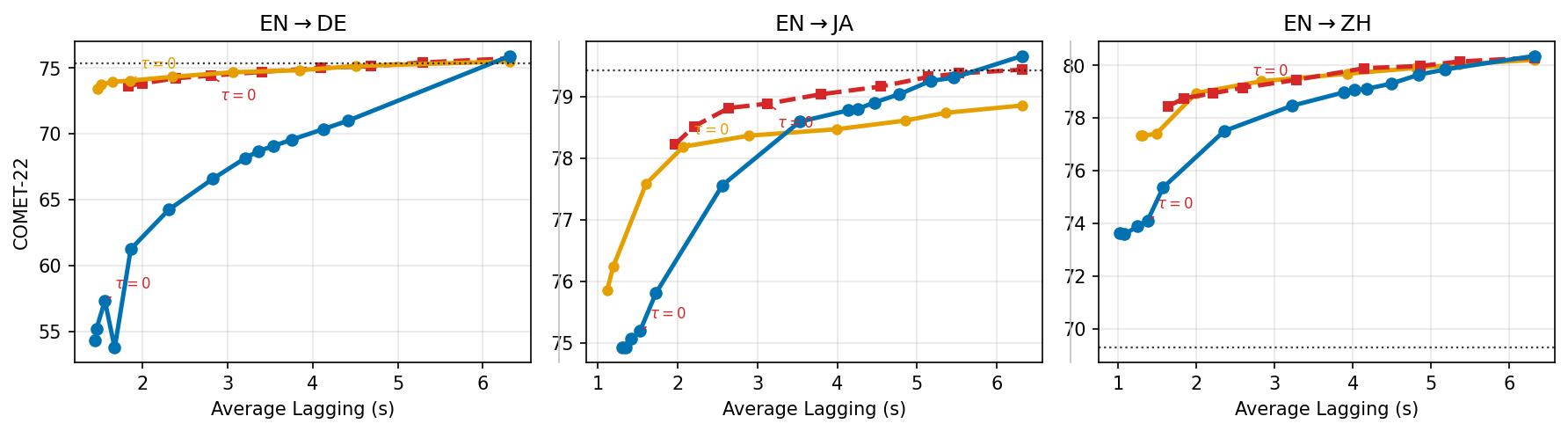}\\
\vspace{2pt}
\includegraphics[width=0.70\textwidth]{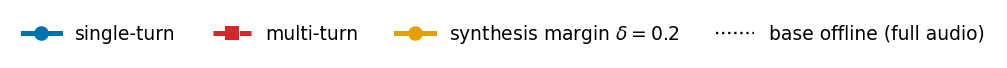}
\caption{Confidence frontiers for the canonical \code{single-turn} and \code{multi-turn} models and the multi-turn $\delta=0.2$ synthesis-margin ablation.}
\label{fig:paper3-speech-staging-representation}
\end{figure*}

Figure~\ref{fig:paper3-speech-staging-representation} and Table~\ref{tab:paper3-speech-staging-representation-auc} compare confidence sweeps for the single-turn, multi-turn, and synthesis-margin $\delta=0.2$ models.

The multi-turn models dominate the frontier over their latency range, especially for EN$\rightarrow$DE/ZH. At the lower latency points, training with a small synthesis margin is better than using the single-turn model, especially for the shorter utterances in CoVoST~2. The single-turn model can achieve even lower latency in some cases, but often at the cost of substantially lower translation quality.

\begin{table}[t]
\centering
\scriptsize
\caption{AUC corresponding to Figure~\ref{fig:paper3-speech-staging-representation}. Each row uses the shared latency interval for that dataset--language panel; best per row in \textbf{bold}.}
\label{tab:paper3-speech-staging-representation-auc}
\setlength{\tabcolsep}{4pt}
\begin{tabular}{@{}lrrr@{}}
\toprule
 & \code{single-turn} & \code{multi-turn} & $\delta=0.2$ \\
\midrule
\multicolumn{4}{@{}l}{\textbf{FLEURS}} \\
EN$\rightarrow$DE & 58.9 & 65.9 & \textbf{67.1} \\
EN$\rightarrow$JA & 80.0 & \textbf{80.1} & 79.2 \\
EN$\rightarrow$ZH & 44.8 & 45.2 & \textbf{45.3} \\
\midrule
\multicolumn{4}{@{}l}{\textbf{CoVoST\,2}} \\
EN$\rightarrow$DE & 67.4 & 72.9 & \textbf{74.5} \\
EN$\rightarrow$JA & 77.8 & \textbf{78.3} & 78.2 \\
EN$\rightarrow$ZH & 77.8 & \textbf{78.9} & 78.8 \\
\bottomrule
\end{tabular}
\end{table}

\subsection{Commit Calibration}

\begin{figure}[!t]
\centering
\includegraphics[width=\linewidth]{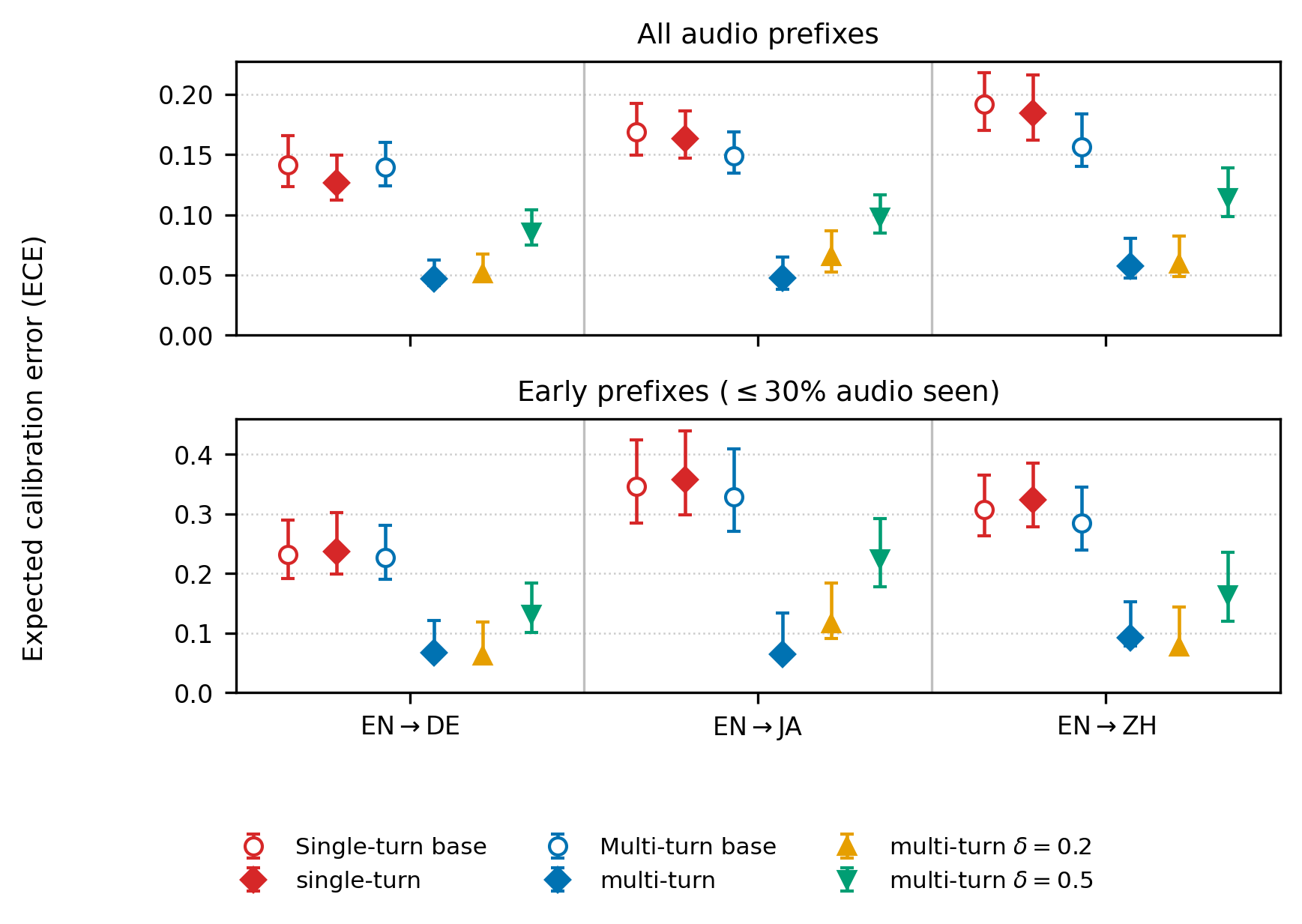}
\caption{Commit-confidence ECE for the single-turn, multi-turn, and synthesis-margin models. Error bars are pointwise $95\%$ utterance-cluster bootstrap intervals.}
\label{fig:paper3-speech-commit-calibration}
\end{figure}

Figure~\ref{fig:paper3-speech-commit-calibration} compares commit calibration for all the models we have presented on 64 base-derived FLEURS oracle chains per direction. Prefix training reduces multi-turn ECE by 63--68\% across all evaluated prefixes, and by 68--80\% on early prefixes. Relative to the matched multi-turn base, $\delta=0.2$ reduces ECE by 56--63\% overall and 65--73\% at early prefixes, while $\delta=0.5$ reduces ECE by 27--39\% overall and 32--42\% at early prefixes.

In contrast, single-turn training reduces overall ECE by only 3--11\%, while early-prefix ECE worsens by 2--5\%.

Early-prefix calibration is especially important because low-latency translation requires the model to decide whether to commit from a context containing little audio.

The multi-turn results show that dense boundary supervision is associated with improved confidence throughout the utterance, which may help explain the stronger low-latency frontier. Calibration alone does not explain the overall differences among models; the quality of prefix and full-sentence supervision may also contribute.

The weak calibration gains of the single-turn model contrast with the text results of \citet{hoang-axelrod-post:2026:wmt}, despite the analogous training objective. We speculate on some of the reasons why this might be the case and why the multi-turn protocol is the better option for speech translation:
\begin{itemize}
\item \textbf{Acoustic segmentation.} Text prefixes end at complete word boundaries and are represented by native language-model tokens. Fixed 2\,s speech blocks instead contain a variable amount of linguistic content, may end within words or phrases, and are encoded by a frozen audio tower. The decoder must therefore infer both the content and completeness of an acoustically uncertain prefix.
\item \textbf{Closing-token semantics.} The closing token may retain its pretrained conversational meaning of ending an assistant response rather than acquiring a clean interpretation as ``wait for more speech,'' weakening the alignment between EOS confidence and translation stability.
\item \textbf{Oracle noise.} The self-teacher oracle is derived from agreement between complete- and partial-waveform translations. Acoustic ambiguity can make these targets noisier than word-bounded text prefixes even after enforcing monotonicity.
\item \textbf{Sequential supervision.} Single-turn training samples independent initial-prefix examples, whereas multi-turn training supervises every boundary in order and preserves the preceding sequence of audio blocks, commitments, and explicit empty turns. This context may expose how evidence accumulates and provide a more direct signal for distinguishing continuation from waiting.
\end{itemize}

%% file: 5_conclusion.tex
\section{Conclusion}

We presented prefix learning for adapting a speech language model to flicker-free simultaneous translation without transcripts or human translations. Training on target prefixes improves the quality--latency frontier for both single-turn forced-prefix and multi-turn decoding. Among the tested inference mechanisms, confidence provides the broadest and most consistently competitive operating range. Under multi-turn training, dense boundary supervision is associated with substantially better confidence calibration across the utterance, including at early prefixes, while single-turn training yields only modest overall gains and no early-prefix improvement. This contrast may help explain why the multi-turn model generally makes better-timed low-latency commitments. A small synthesis margin sometimes extends low-latency coverage, particularly on shorter utterances, whereas a larger margin degrades translation quality and calibration. Prefix adaptation therefore improves simultaneous speech translation, especially under multi-turn append-only decoding, while synthesis density introduces a non-monotonic quality--latency trade-off.

%% file: 6_limitations.tex
\section{Limitations}
\begin{itemize}
\item Experiments cover one base model (Qwen2.5-Omni-7B), three English-source language directions, and read or prompted speech benchmarks. Transfer to other model families, languages, accents, domains, and spontaneous conversational speech remains untested.
\item Stable-prefix targets are synthesized by the base model using lexical longest-common-prefix agreement with its own full-audio translation. They can inherit self-teacher errors and biases, and lexical stability does not guarantee semantic support or correctness.
\item Source time advances in fixed $2$\,s blocks with known utterance boundaries. We do not evaluate adaptive segmentation, continuous unsegmented input, network delay, wall-clock decoding latency, throughput, or memory use; Average Lagging measures policy delay only, and the expected computational benefit of key--value cache reuse is not quantified.
\item Translation quality is measured automatically with COMET, without human evaluation of adequacy, fluency, or the usefulness of partial outputs. Output is monotone by construction, so the system prevents flicker but cannot revise an early erroneous commitment.
\item Calibration is evaluated on 64 FLEURS oracle chains per direction against base-derived continue-versus-stop labels. The results characterize commit confidence for these oracle commit-decision contexts and prompt representations, not generic language-model calibration, human judgments of semantic certainty, or out-of-domain calibration.
\item The single-turn and multi-turn comparisons change both the causal representation and the structure of training examples. Without a crossed train--inference representation study, their differences cannot be attributed solely to matching commit-decision contexts.
\item We did not audit Qwen2.5-Omni's pretraining data for overlap with the evaluation sets, so benchmark contamination cannot be ruled out.
\end{itemize}

%% file: 7_appendix.tex
\section*{Appendices}

\section{Reproducibility Details}
\label{sec:reproducibility}

\paragraph{Audio and target units.} Audio is resampled to 16\,kHz and segmented at the fixed $2$\,s cadence dictated by Qwen2.5-Omni's audio implementation; this interval is not a tuned inference-mechanism parameter. A final partial block is retained unless it is shorter than $0.5$\,s, in which case it is merged with the preceding block. German target prefixes are segmented into whitespace-delimited words; Japanese and Chinese use characters. Prefix synthesis, training, validation, and evaluation use the same language-specific target unit.

\paragraph{Corpus and splits.} The assembled mixture contains $139{,}684$ distinct English utterances: $27{,}937$ LibriSpeech-100, $69{,}842$ LibriSpeech-360, and $41{,}905$ Common Voice utterances. The same utterances and block boundaries are reused for all three target languages and both representations. Every 50th chain in the assembled mixture is withheld from optimization; controlled selection uses at most the first 400 held-out chains. A post-selection EN$\rightarrow$ZH diagnostic uses all 887 held-out Common Voice chains. Final evaluation uses 585 FLEURS utterances and 1,999 usable CoVoST~2 utterances per language direction.

\begin{table}[H]
\centering
\scriptsize
\caption{Training, validation, and test data.}
\label{tab:data-splits}
\setlength{\tabcolsep}{2pt}
\begin{tabular}{@{}p{0.33\columnwidth}p{0.18\columnwidth}rrr@{}}
\toprule
Dataset / genre & Reference & $N$ & \shortstack{Train\\mix} & \shortstack{Avg.\\audio (s)} \\
\midrule
\multicolumn{5}{@{}l}{\textbf{Training}} \\
LibriSpeech-100\newline\emph{Audiobook speech} & Self-teacher & $27{,}403$ & $20.0\%$ & $12.7$ \\
LibriSpeech-360\newline\emph{Audiobook speech} & Self-teacher & $68{,}469$ & $50.0\%$ & $12.6$ \\
Common Voice\newline\emph{Crowd-sourced read speech} & Self-teacher & $41{,}018$ & $30.0\%$ & $5.7$ \\
\cmidrule(lr){1-5}
\multicolumn{2}{@{}l}{\textbf{Training total}} & $136{,}890$ & $100.0\%$ & --- \\
\midrule
\multicolumn{5}{@{}l}{\textbf{Validation}} \\
Mixed holdout\newline\emph{Audiobook + crowd-sourced read} & Self-teacher & $400$ & --- & $10.4$ \\
LibriSpeech-100 holdout\newline\emph{Audiobook speech} & Self-teacher & $400$ & --- & $13.0$ \\
FLEURS development\newline\emph{Multilingual read speech} & Human & $347$ & --- & $9.4$ \\
\midrule
\multicolumn{5}{@{}l}{\textbf{Test}} \\
FLEURS test\newline\emph{Multilingual read speech} & Human & $585$ & --- & $9.7$ \\
CoVoST~2\newline\emph{Crowd-sourced read speech} & Human & $1{,}999$ & --- & $5.8$ \\
\bottomrule
\end{tabular}
\end{table}

\paragraph{Optimization.} All primary adapters start from Qwen2.5-Omni-7B. The audio and vision towers remain frozen. LoRA is applied only to the text decoder's attention $q$, $k$, $v$, and output projections and MLP gate, up, and down projections, with rank 32, scaling factor 64, dropout $0.05$, and no bias. Training uses bfloat16, AdamW, a constant learning rate of $5\times10^{-6}$, and an effective batch size of 64. Multi-turn records longer than 1,200 tokens are omitted rather than truncated. The random seed is 42. Checkpoints are retained every 500 updates through the 6,000-update ceiling.

\begin{table}[t]
\centering
\small
\begin{tabular}{@{}lcc@{}}
\toprule
Direction & \code{single-turn} & \code{multi-turn} \\
\midrule
EN$\rightarrow$DE & 2,000 / 6,000 (384k) & 3,000 / 6,000 (384k) \\
EN$\rightarrow$JA & 2,500 / 6,000 (384k) & 5,500 / 6,000 (384k) \\
EN$\rightarrow$ZH & 1,000 / 6,000 (384k) & 3,000 / 3,000 (192k) \\
\bottomrule
\end{tabular}
\caption{Selected update / final update evaluated; parentheses give the total number of optimized records through the final update (updates multiplied by effective batch size 64).}
\label{tab:paper3-training-details}
\end{table}

\paragraph{Average Lagging.} Let an utterance contain $N$ source blocks and let the final output contain $T$ target units. If $g_i$ is the number of source blocks read when target unit $i$ is committed, $\gamma=T/N$, and $\tau$ is the first target position emitted after all $N$ blocks have been read (or $T$ if this never occurs), we compute
\begin{equation}\label{eq:average-lagging}
\begin{aligned}
\operatorname{AL}_{\mathrm{blocks}}&=\frac{1}{\tau}\sum_{i=1}^{\tau}\left(g_i-\frac{i-1}{\gamma}\right),\\
\operatorname{AL}_{\mathrm{sec}}&=2\operatorname{AL}_{\mathrm{blocks}}.
\end{aligned}
\end{equation}
An empty output is assigned source-end latency. Frontier selection uses attainable COMET: at each latency budget, a candidate receives the best COMET among its operating points at or below that budget, and these values are averaged over a common grid.

\paragraph{Attainable-envelope AUC.} Each dataset--language panel uses one shared latency interval $[L,U]$, where $L$ and $U$ are respectively the minimum and maximum Average Lagging in seconds over the union of all operating points from all compared sweeps in that panel. We evaluate 101 equally spaced budgets $b_j=L+j(U-L)/100$. Let $S_s$ be sweep $s$'s set of (Average Lagging, COMET) points, let $q_{\mathrm{worst}}$ be the lowest COMET value anywhere in the panel, and define
\begin{equation}\label{eq:attainable-envelope-auc}
\begin{aligned}
\mathcal{Q}_s(b)&=\{q:(a,q)\in S_s,\ a\le b\},\\
F_s(b)&=\begin{cases}
\max \mathcal{Q}_s(b),&\mathcal{Q}_s(b)\ne\varnothing,\\
q_{\mathrm{worst}},&\text{otherwise},
\end{cases}\\
\operatorname{AUC}_s&=\frac{1}{101}\sum_{j=0}^{100}F_s(b_j).
\end{aligned}
\end{equation}
Thus every sweep has the same domain and denominator, including below its minimum attainable latency. Tables~\ref{tab:paper3-speech-staging-auc} and~\ref{tab:paper3-speech-frontier-il-auc} compare the base and selected trained models under wait-$k$, target suffix deletion, and confidence using one shared interval across all six sweeps in each panel; Table~\ref{tab:paper3-speech-staging-representation-auc} applies the same calculation to the three confidence frontiers in Figure~\ref{fig:paper3-speech-staging-representation}.

\paragraph{Cached decoding.} Both representations exploit the block-local audio encoder and retain the fixed prefix and processed audio across boundaries. In the single-turn forced-prefix protocol, the committed target and newly generated delta occupy a temporary cache extension that is cropped after decoding; the growing committed target must therefore be replayed at the next boundary. The multi-turn append-only cache is never cropped: each audio block and accepted assistant delta is appended permanently, so prior target states are reused as well as prior audio states. The single-turn forced-prefix and multi-turn append-only protocols apply the same confidence, wait-$k$, and suffix-deletion mechanisms within their respective prompt layouts. We do not report wall-clock latency, throughput, or memory use, so this computational motivation should not be interpreted as an empirical efficiency comparison.

\section{Translation Prefix Synthesis}
\label{sec:stable-prefix-synthesis-algorithm}

Algorithm~\ref{alg:stable-prefix-synthesis} constructs one training chain from an utterance. The normalization step removes surrounding whitespace and collapses internal whitespace before target-unit segmentation. The full-audio decode uses the final prompt containing the end-of-audio marker; every proper prefix uses the streaming prompt. At each partial-audio decoding step, the aligned full-audio token replaces the greedy token when it is within synthesis margin $\delta$ of the greedy probability. Setting $\delta=0$ recovers ordinary greedy prefix decoding. After decoding, lexical agreement and its running maximum determine the monotone target commitment; the final boundary commits the complete anchor regardless of earlier agreement.

\begin{algorithm}[t]
\caption{Self-anchored translation-prefix synthesis}
\label{alg:stable-prefix-synthesis}
\begin{algorithmic}[1]
\Require waveform $a$; boundaries $0<t_1<\cdots<t_K=|a|$; frozen model $g$; synthesis margin $\delta\in[0,1]$; generation limit $L$; model-token functions $\operatorname{tok}$ and $\operatorname{detok}$; target-unit functions $\operatorname{seg}$ and $\operatorname{join}$
\Ensure monotone commitments $c_1,\ldots,c_K$
\State $f \gets \operatorname{normalize}(g(a;\ \textsc{final}))$
\State $F \gets \operatorname{seg}(f)$
\State $m \gets 0$
\For{$i \gets 1$ to $K-1$}
    \State $R_i \gets \operatorname{tok}(f\circ\textsc{eos};\ a_{\leq t_i},\ \textsc{streaming})$
    \State $H_i \gets ()$
    \While{$|H_i|<L$}
        \State $p \gets g(\cdot\mid a_{\leq t_i},H_i;\ \textsc{streaming})$
        \State $x \gets \arg\max_v p(v)$; $q \gets |H_i|+1$
        \If{$q\leq |R_i|$}
            \State $r \gets R_{i,q}$
            \State $y \gets r$ if $x=r$ or $p(r)+\delta>p(x)$, otherwise $y \gets x$
        \Else
            \State $y \gets x$
        \EndIf
        \State $H_i \gets H_i\circ y$
        \If{$y=\textsc{eos}$}
            \State \textbf{break}
        \EndIf
    \EndWhile
    \State $h_i \gets \operatorname{normalize}(\operatorname{detok}(H_i))$
    \State $\ell_i \gets \left|\operatorname{lcp}(\operatorname{seg}(h_i),F)\right|$
    \State $m \gets \max(m,\ell_i)$
    \State $c_i \gets \operatorname{join}(F_{1:m})$
\EndFor
\State $c_K \gets f$
\State \Return $(c_1,\ldots,c_K)$
\end{algorithmic}
\end{algorithm}

\section{Chinese Self-Teacher Cleanup}
\label{sec:zh-anchor-cleaning}

Inspection of the EN$\rightarrow$ZH self-teacher outputs revealed trailing assistant-style clauses that invited the user to provide more audio, ask another translation question, or request help. The cleaner considers only suffixes beginning at a Chinese or ASCII sentence, clause, or comma boundary. It removes a suffix when the suffix begins with a conditional expression such as ``if there is more'' and contains both a topic cue (audio, content, translation, question, or English learning) within its first 80 characters and an invitation cue (tell me, ask me, contact me, continue, anytime, welcome, or help) within its first 120 characters. It also removes direct suffixes beginning with an equivalent of ``you can continue'' or ``you can tell me anytime.'' The rule is applied to a fixed point so that nested boilerplate is removed, while punctuation and closing quotation marks belonging to the retained translation are preserved.

After cleaning the complete-audio anchor, every intermediate commitment is truncated to the cleaned anchor length. The reconstruction procedure verifies that every row remains a prefix of the cleaned anchor, commitments remain monotone, and the final row equals the cleaned anchor. Chains whose anchors contain no translation after cleaning are quarantined rather than converted into empty training targets.

The audit covers all 139,684 EN$\rightarrow$ZH chains. The procedure changes 92,039 chains (65.89\%): 36,684 of 41,905 Common Voice chains and 55,355 of 97,779 LibriSpeech chains. Eight boilerplate-only chains are quarantined. Manual review covers 20 sampled changed chains and all 15 chains flagged as ambiguous or empty after cleaning. Seven ambiguous cases contain legitimate uses of expressions resembling the invitation cues and are conservatively left unchanged; the remaining eight are the quarantined boilerplate-only chains.

\section{LLM Prompt Templates}
\label{sec:llm-prompt-templates}

Both contracts use Qwen2.5-Omni's native system persona and one fixed streaming instruction. The instruction is not rewritten at the final boundary; the literal marker \code{[end of audio]} is appended after the final audio block. The examples below show EN$\rightarrow$DE. For Japanese and Chinese, the language name changes and no whitespace is inserted after a forced target prefix.

\subsection{Shared System Message and Instruction}

{\scriptsize
\begin{verbatim}
[system] You are Qwen, a virtual human
         developed by the Qwen Team, Alibaba
         Group, capable of perceiving auditory
         and visual inputs, as well as
         generating text and speech.

[instruction] You are translating ongoing
         English speech into German in real
         time. You will hear the speech in
         consecutive audio segments. After each
         segment, output ONLY the additional
         German translation that has become
         certain from the audio so far, and
         nothing else; if nothing new is yet
         certain, output nothing. When you see
         [end of audio], output the remaining
         translation needed to complete the
         utterance.
\end{verbatim}
}

\subsection{Single-Turn Forced-Prefix Decoding Template}
\label{sec:single-turn-template}

Each decision is rendered as a single user turn containing the complete audio observed so far. The previously committed translation is supplied as an open assistant prefix; only the continuation and closing token are generated and supervised. The initial partial and initially complete training tasks leave \code{[forced]} empty.

{\scriptsize
\begin{verbatim}
[system]    <Qwen2.5-Omni system persona>
[user]      <fixed instruction> <audio prefix>
[forced]    Wissenschaftler entdeckten
-->         eine neue Art

[system]    <Qwen2.5-Omni system persona>
[user]      <fixed instruction>
            <complete audio> [end of audio]
[forced]    Wissenschaftler entdeckten
            eine neue Art
-->         im Regenwald.
\end{verbatim}
}

At inference, the system message, instruction, and audio prefix remain in the persistent cache. The forced and generated target region is recomputed at each boundary and then discarded from the cache, while the visible target commitment remains unchanged.

\subsection{Multi-Turn Append-Only Decoding Template}
\label{sec:multi-turn-template}

The fixed instruction appears only with the first audio block. Subsequent user turns append one new block, and each assistant turn contains only the newly committed delta. An empty assistant turn is an explicit wait.

{\scriptsize
\begin{verbatim}
[system]    <Qwen2.5-Omni system persona>
[user]      <fixed instruction> <audio block 1>
[assistant] Wissenschaftler
[user]      <audio block 2>
[assistant] entdeckten eine neue
            Art
[user]      <audio block 3> [end of audio]
[assistant] im Regenwald.
\end{verbatim}
}

Training masks the system, user, instruction, and audio-placeholder tokens and supervises every assistant delta plus its closing token. During inference the entire sequence remains in one persistent cache, so each decision can attend to all earlier audio blocks and committed deltas but never to future audio.